# 4D flight trajectory prediction using a hybrid Deep Learning prediction method based on ADS-B technology: a case study of Hartsfield–Jackson Atlanta International Airport (ATL)

Hesam Sahfienya, Islamic Azad University Science and Research Branch, hesam.shafienya@srbiau.ac.ir

Amelia C. Regan, Department of Computer Science, University of California, Irvine, aregan@uci.ed (corresponding author)

**Abstract**

*The core of any flight schedule is the trajectories. In particular, 4D trajectories are the most crucial component for flight attribute prediction. Each trajectory contains spatial and temporal features that are associated with uncertainties that make the prediction process complex. Today because of the increasing demand for air transportation, it is compulsory for airports and airlines to have an optimized schedule to use all of the airport's infrastructure potential. This is possible using advanced trajectory prediction methods. This paper proposes a novel hybrid deep learning model to extract the spatial and temporal features considering the uncertainty of the prediction model for Hartsfield–Jackson Atlanta International Airport (ATL). Automatic Dependent Surveillance-Broadcast (ADS–B) data are used as input to the models. This research is conducted in three steps: (a) data preprocessing; (b) prediction by a hybrid Convolutional Neural Network and Gated Recurrent Unit (CNN-GRU) along with a 3D-CNN model; (c) The third and last step is the comparison of the model's performance with the proposed model by comparing the experimental results. The deep model uncertainty is considered using the Mont-Carlo dropout (MC-Dropout). Mont-Carlo dropouts are added to the network layers to enhance the model's prediction performance by a robust approach of switching off between different neurons. The results show that the proposed model has low error measurements compared to the other models (i.e., 3D CNN, CNN-GRU). The model with MC-dropout reduces the error further by an average of 21 %.*

## 1. Introduction

To be sustainable, airports and airlines must use all of their infrastructure potential to optimize their activities and maximize their benefits as the demand for air transportation increases. Major airports need optimized flight schedules with millions of aircraft trajectories. The Statistica website shows an increase in passenger traffic and flight operation between 2000 and 2019, from nearly 80 to 110 million passengers, demonstrating the importance of this research [1].
   The United States follows the Next Generation Air Transportation System (NextGen) program and the Euro control called Single European Sky ATM Research (SESAR) [2-4]. These two programs and the feature of china's Next Generation of Air Traffic management program are Trajectory-Based Operations (TBO). It shows the importance of trajectories and trajectory predictions in air traffic management. This paper focuses on the NextGen program based on the selected case study, which aims to optimize the flight schedule and make the flights safe by helping the controllers handle flights.

The world's busiest airport, Hartsfield–Jackson Atlanta International Airport (ATL), is selected as the case study of this research. Hartsfield-Jackson earned the world's busiest airport title after seeing more than 107 million passengers in 2018 and held that title for 21 consecutive years. It is important to optimize the usage of infrastructure in this airport and its sky for this airport and operational airlines and all over the world because of the consecutive effects of world flights on each other.

4D trajectory prediction includes time, longitude, latitude, and altitude. It calculates these four main aircraft parameters for the future waypoint sequence based on the historical flight data. According to the time scale, 4D trajectory prediction can be divided into the following categories [5]:

1. **Tactical (short-term) prediction:** Short-term approaches require the fastest response to handle emergencies on the way, which will be within several minutes or even shorter.
2. **Strategical (long-term) prediction:** Long-term approaches can take advantage of full knowledge to predict trajectories strategically and gives a reliable perspective

We use strategic 4D trajectory prediction, and this prediction can be affected by uncertainties such as pilot bias, wind, weather conditions, etc. These uncertainties could decrease our model's prediction efficiency. Accordingly, we propose a dropout method based on a Bayesian model to manage the uncertainty problem and increase the efficiency of the proposed prediction model. The deep learning approach used in this paper has been employed to process time series-related predictions such as pedestrian trajectory, vehicle trajectory, traffic flow prediction, etc., in the past [5,6].

This paper proposes a hybrid CNN-GRU with a C3D Deep learning model for strategical (long-term) 4D trajectory prediction for ATL airport. CNN's automated deep feature extraction power efficiently and accurately predicts the trajectory's spatial and temporal separated data. This approach contains four main parts. In the first step, (a) the ADS-B dataset is prepared by interpolating the data and then separating the dataset into spatial and temporal parts; (b) Principal Component Analysis (PCA) is implemented with dimensionality reduction purpose; (c) sliding window technique is used for input generation to implement the proposed hybrid prediction model to obtain the results (d) the results of the proposed model and a comparison between the proposed model and others deep learning models are presented. We demonstrate the best model for strategic flight trajectories prediction in ATL airport and its sky.

This paper is organized as follows. In section 2, we introduce the related works. Section 3 describes the opensky data that provides the foundation for our study. In section 4, we describe our approach and methodology in detail. In section 5, we share the results of our experimental analysis. We conclude in section 6.

## 2. Related works

With the progression of surveillance equipment, being the compulsory and necessary part of the aircraft, increasing worldwide use of this technology, and the advantage of easy access to data, usage of them is increasing dramatically. According to this, the 4D trajectory prediction, assessing the schedule, and optimization field obtained so much attention from the air traffic organizations and researchers worldwide. Zhang, $et\ al$. [8] proposed an online four-dimensional trajectory prediction (4D-TP) method using an ADS-B Receiver and corresponding data processing algorithm. That paper also presents an exceptionally clear tutorial on this problem. Zhang, $et\ al$. [9] increased en-route flight safety by developing deep learning models for trajectory prediction, where model prediction uncertainty is characterized by a Bayesian approach applied through a hybrid deep learning model. Wang, $et\ al$. [10] proposes a novel hybrid model to address the short-term trajectory prediction problem in Terminal Maneuvering Area (TMA) by applying machine learning methods. This paper uses Multi Convolutional Neural Network (MCNN) and Random Forest (RF) models and, in the end, compares MAE and RMSE errors. Ma, $et\ al$. [11] propose a 4D trajectory prediction hybrid architecture based on deep learning, which combined Convolutional Neural Network (CNN) and Long Short-Term Memory (LSTM). According to this research, this hybrid approach has more accuracy than implementing CNN and LSTM networks. Guo, *et al*. [12] proposes an innovative deep learning-based mapping to cube architecture for network-wide urban traffic forecasting to improve the limitation of traffic data availability and quality. Experiments using real taxi GPS vehicle trajectory data confirm the accuracy and effectiveness of the proposed approach combining 3-Dimensional Convolutional Networks (C3D) with Convolutional Neural Networks (CNNs) and Recurrent Neural Networks (RNNs), called CRC3D as a hybrid method integrating CNN-LSTM and C3Ds. Lambeth, *et al*. [13] propose a generic assessment of such strategic schedules using prediction about arrival/departure flight delays and cancellations for London Heathrow Airport (LHR). Shi, *et al*. [14] propose a constrained Long Short-Term Memory network for flight trajectory prediction. Wu, *et al*. [15] solved the problem that traditional trajectory prediction methods cannot meet the requirements of high-precision, multi-dimensional, and real-time prediction model based on backpropagation (BP) neural network was studied and trained to utilize Automatic Dependent Surveillance-Broadcast (ADS-B) trajectory from Qingdao to Beijing to predict the flight trajectory at future moments. Zeh, *et al*. [16] proposed an approach applied to typical uncertainties in trajectory prediction, such as the actual take-off mass, non-constant true airspeed, and uncertain weather conditions. In general, this paper's approach applies to all sources of quantifiable interdependent uncertainties. The vast use of hybrid models and their combination with ADS-B Surveillance data is clear, and scientists worldwide are interested in this field because of its advantages and flight-related organization's demand for Trajectory-based operations programs such as NextGen and SESAR.

**3. Data**

Automatic Dependent Surveillance-Broadcast (ADS-B) is a technology to connect aircraft and ground stations consecutively. This connection is of two types: (a) Aircraft-Aircraft; (b) Aircraft-Ground Stations. ADS-B data provides the vast majority of data about aircraft, signals, sensors, etc. ADS-B messages consist of static and dynamic components. The static data implies the taking-off and landing airports, callsign, and scheduled airways. The dynamic component is 4D data, which mainly includes timestamps, positions (i.e., latitude, longitude, and altitude), velocity, and heading [13]. The dynamic elements of the ADS-B data type are used in this paper.

The dataset used in this paper is taken from the OpenSky network via limited access to the data set. We mine the trajectory data for ATL airport, historical data from March 30, 2016, to March 30, 2020. Each trajectory shows aircraft taking-off, landing, and flying over ATL airport, and every trajectory contains the data mentioned in table 1.

**Table 1.** Features of trajectories

| Features | Unit | Instance in trajectory |
|---|---|---|
| Timestamp | Unix | 1478874138 |
| Aircraft Number | Icao24 | aaa83f |
| Geographical Information | Degree | 33.79832 |
| Velocity | Knot | 221.5576 |
| Heading | Degree | 348.4813 |
| Vertical Rate | Feet/Min | -0.32512 |
| Call sign | XXX | EJA786 |
| Hour | Unix | 1478872800 |

Each data record with the same number belongs to the aircraft $i$, and collection of all that records form the aircraft as presented as the trajectory $T_i$ [17]. In this paper, we use all trajectory data from the geographic positions of ATL airport and take advantage of trajectory data for flight departure, arrival, and flying over the ATL airport.

## 4. Methodology

This section proposes a hybrid deep learning model for predicting the 4D trajectory of flights through historical ADS-B data. The approach includes two main models:
  1. A hybrid CNN-GRU model to fully extract spatial-temporal features of flight trajectories.
  2. A three-dimensional CNN to predict spatial-temporal feature flight state.

After extraction of the dataset into these two different parts, we congregate the results and merge them to generate the high-level and accurate spatial-temporal feature mapping. In the last layer of the proposed regressor model, the linear activation function is implemented to predict 4D flight trajectories. In this paper, we characterize network model prediction uncertainty from a Bayesian perspective utilizing the Monte Carlo Dropout method. Model prediction uncertainty is quantified with the dropout strategy and propagated through the integrated model with Monte Carlo samples [8].

The proposed architecture of the 4D trajectory prediction utilizing deep hybrid CNN-GRU model based on ADS-B data is shown in Fig. 1. As shown in Figure 1, ADS-B signals broadcast from the airplanes to the ground stations. After storing in the cloud center at Opensky, we preprocess the data and predict the trajectory values through a deep CG3D model.

In section 3.1, the ADS-B data are preprocessed. In this part, the missing data are imputed by means of cubic spline interpolation in the dataset. Section 3.2 separates the data into two separate groups: (a) Spatial; (b) Temporal. Section 3.3 presents the prediction of the 4D flight trajectories propagated by the sliding window method for ATL airport using the proposed CNN-GRU with a 3D CNN approach. 3.4 model prediction uncertainty is characterized from a Bayesian perspective through Monte Carlo and compares the model with MC Dropout and the model without MC Dropout.

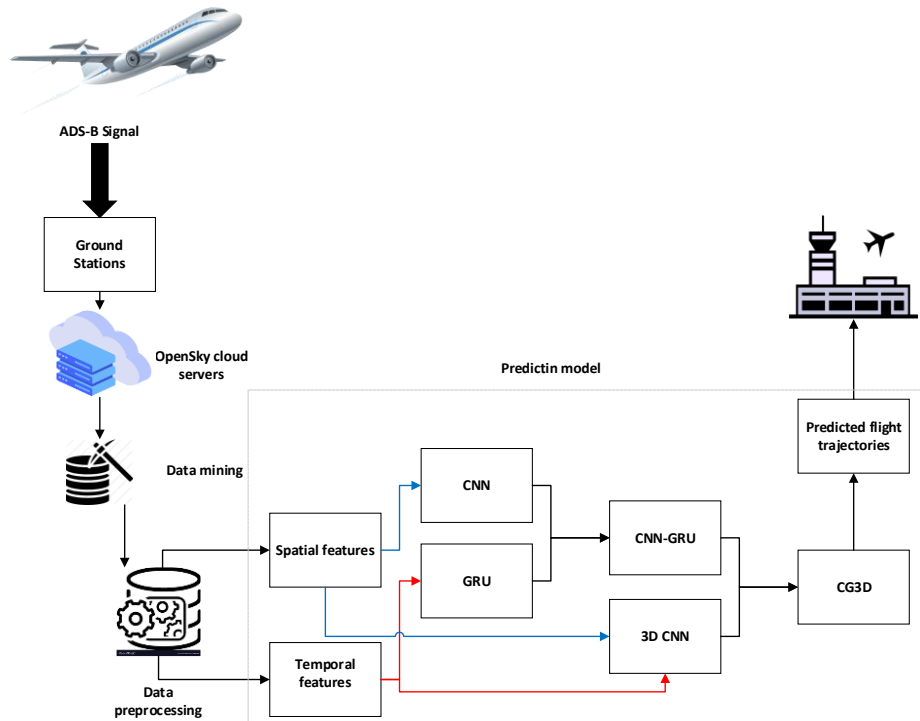

**Figure 1.** Methodology schematic

## 4.1 Data Preprocessing

Due to system errors, signal occlusion, etc., the real ADS-B trajectory data has problems such as repeated trajectory samples and missing trajectory samples. These problems in the ADS-B dataset are common due to interrupted surveillance of the data between the airplanes and ground receivers [10]. These data anomalies affect the prediction performance, and the trajectories with a large amount of missing data need to be preprocessed. The interpolation methods can be used for this aim.

In this part, we prepare the dataset for the next sections and, particularly, prediction. First, we interpolate the data to take care of missing data using cubic spline interpolation, and this method interpolates the dataset smoothly and more accurately. A sliding window approach contains 100 trajectory data points and slides over every five trajectories to generate the input and output data to feed the network. Principal Component Analysis (PCA) is applied to the separated data to scale and reduce the dimensions of the data. The schema of the preprocessing progress is shown in Fig. 2.

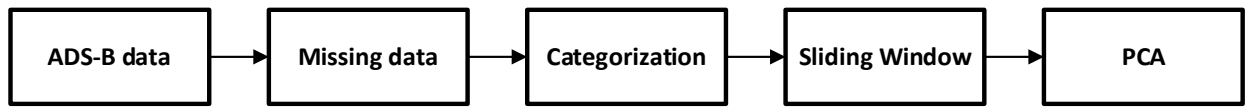

**Figure 2.** Preprocessing the dataset

## 4.2 Categorize the dataset

According to the different parts of the data variables and the need to extract them with other types, we divide them into two separate categories: spatial and temporal. Spatial data contains attributes such as longitude, latitude, and altitude, etc. And also, temporal data contains attributes such as time, velocity, etc. The separation of the input data with special features can improve the performance of different models.

Feeding the CNN with spatial data and GRU with temporal enhance the prediction efficiency. GRU has a low memory structure, and because of that, it is appropriate for temporal feature extraction. The GRUs are the best for short-term prediction, and to have a robust model for long-term strategical prediction, and this model needs to be combined with CNN to handle the spatial features.

## 4.3 Prediction

4D Predicted trajectories are the principal elements of this approach. This section proposes a deep hybrid model with high performance in training data. This hybrid model contains a combination of a hybrid Convolution Neural Network (CNN) and sub-model of Recurrent Neural Network (RNN), called Gated Recurrent Units (GRU) with a 3-Dimensional Convolution (C3D), called CG3D, which use is for 4D trajectory prediction. CNN-GRU is a powerful deep learning model, a CNN part for spatial data extraction and GRU for long-term temporal data extraction. In this approach, the C3D model is used to predict future flight trajectory states. Each model is fed with appropriate spatial or temporal features and then assemble the results to generate a high-level spatiotemporal feature mapping. Next step, according to the results, we implement the Monte Carlo (MC) Dropout method for uncertainty and compare it with the model MC

Dropout uncertainty. We put the features into a linear activation in the last part to obtain the best prediction performance.

**4.3.1 Sliding window**

This subsection's purpose is to define a sliding window along the dataset trajectory vectors timestamps to generate the input data. The ADS-B Dataset used in this paper has millions of time series with multi-dimensions that correspond to the time series. The principle is: first, we define a window consisting of 100-time series, and each of them has the dimensions of the trajectory, icao24, longitude, latitude, altitude, velocities. This window takes the 100 timestamps as the training data. It predicts the next five timestamps, and then in the next step, the window slides down on the dataset. It continues to where the last trajectory predicted [18]. The whole process of the sliding window is shown in Fig. 3. We should accomplish this step as a step-up preprocessing to do preprocessing on the dataset.

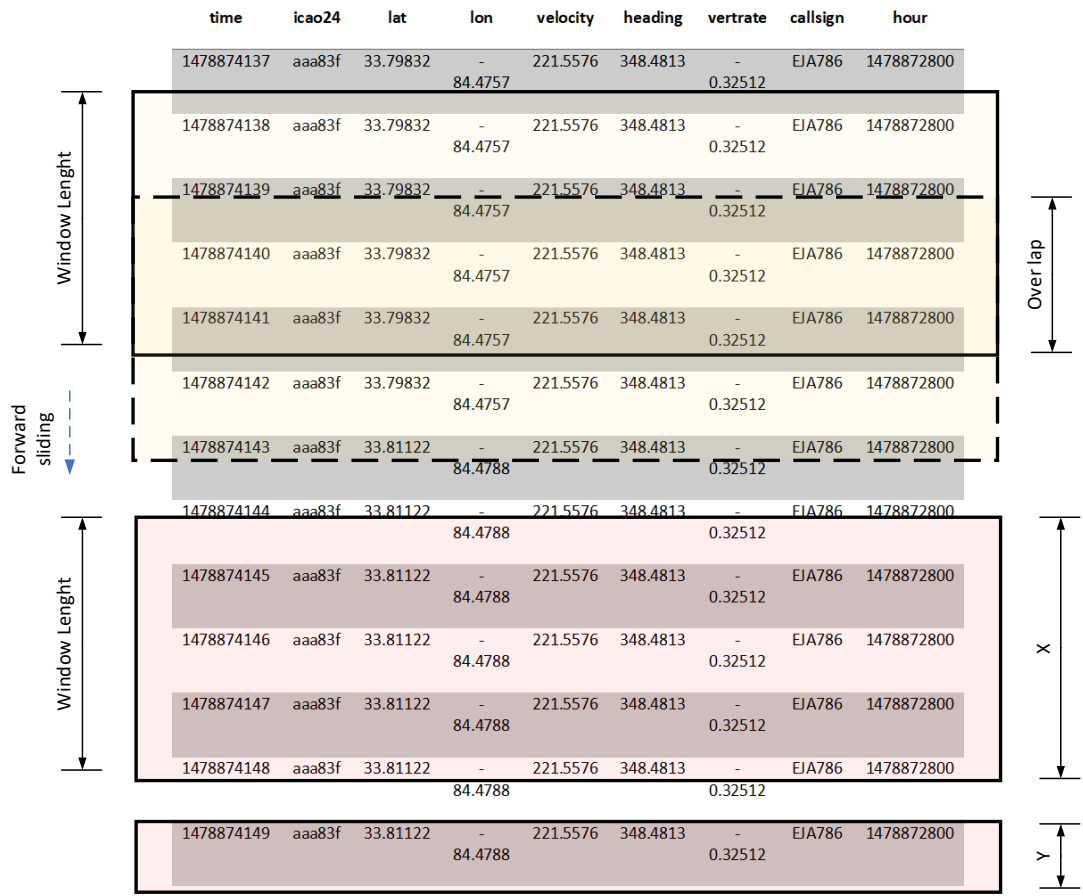

**Figure 3.** Sliding Window Mechanism

**4.3.2 CNN-GRU**

As mentioned before, one of the two parallel parts of our hybrid approach is CNN-GRU. This hybrid model extracts the Spatio-temporal features of the trajectories.

CNN is a neural network with a deep structure. Every CNN consists of four main layers, data matrix input, convolution, pooling, and fully connected layer. This method is powerful in processing image-related problems [19-20]. In this paper, we take advantage of this attribute and use it for spatial extraction in our model. CNN is convolution, and the difference between convolution and fully connected convolutional performance takes full advantage of the information in the nearby areas of the data matrix. Parameter matrix size reduction is implemented by means of sparse connection and sharing weights. Next step, this model builds a feature map with a pooling layer. We could build a CNN containing more than one convolution and pooling layer. More layers of convolution and pooling, and we could get more accurate predictions to extract the spatial features, and by means of this built series, we could actually extract the spatial features in every time frame [21]. We take advantage of CNN to extract the spatial features of the trajectory data. The structure of the CNN model is shown in Fig. 4.

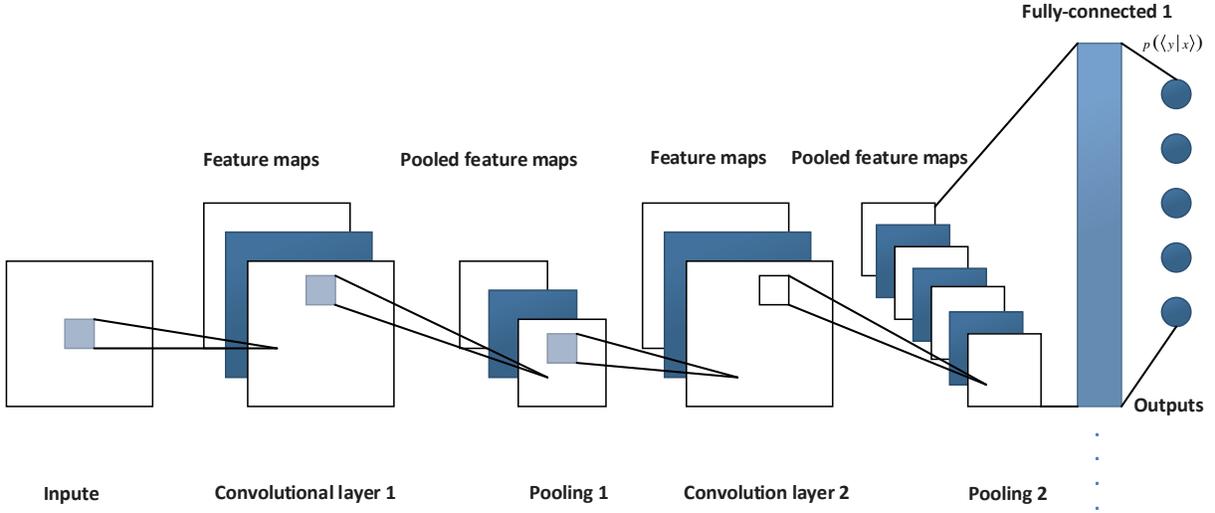

**Figure 4.** CNN Architecture

In terms of mathematic, we set $O_{mn}^{xy}$ As a value of the cell at the position (x, y) in the $nth$ feature map and $mth$ layer, it is achieved from equation (1).

$$O_{mn}^{xy} = \tanh(b_{mn} + \sum_h \sum_{p=0}^{P_m-1} \sum_{q=0}^{Q_m-1} \omega_{mnh}^{pq} \cdot O_{(m-1)m\backslash h}^{(x+p)(y+q)}) \quad (1)$$

Where $f(.)$ is the activation function .e.g., ReLU (Rectifier Linear Unit). $b_{mn}$ shows bias and $\omega_{mnh}^{pq}$ is weight, where $h$ is the index of the feature map in the $(m-1)$ layer and $(p,q)$ defines the position of the weights in the filter. $P_m$ and $Q_m$ demonstrate respectively the height and width of the filter. By means of contiguous calculations, we reduce the size of feature maps. So, the CNN output is a fully-connected layer.

GRU is a derived method from the RNN, which is appropriate for processing time series and extracting the temporal features. During backpropagation, RNN has the problem of gradient vanishing. Gradients are values that are used to update a neural network's weights. It has an internal mechanism called a gate, which can regulate the flow of information.

LSTM is another alternative for solving the short memory problem of the RNN, which has all the advantages of the RNN plus the ability to solve the long-term dependencies problem without vanishing the gradient effect, just like the GRU. This method is implemented for long-term trajectory data prediction. This method works the same as the GRU. The whole mechanism and difference between them are shown in Fig. 4. [22].

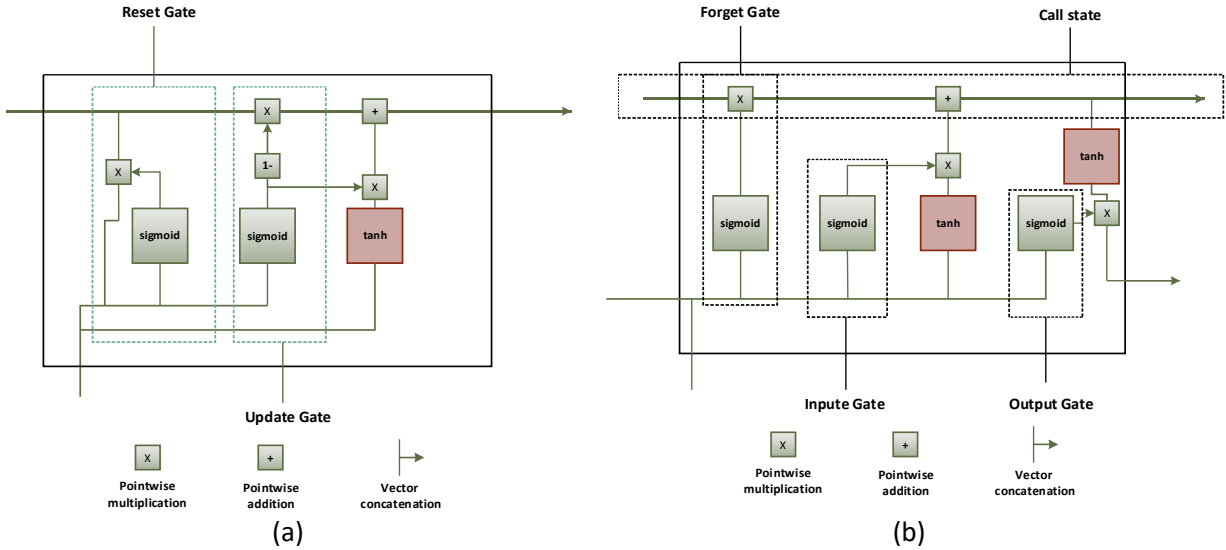

**Figure 5.** (a) GRU Structure; (b) LSTM Structure

The mechanism in the LSTM is to add a cell layer to the model to make sure the transfer of hidden state from one iteration to the next is appropriate and reasonable. Gated Recurrent Unit (GRU) takes everything related to this subject to a higher level, and the GRU has gating units that adjust the information flow inside the unit without dedicating an individual memory cell. Thus, GRU can efficiently be used for long-term temporal predictions. According to the higher level of GRU and its efficiency, we discuss the mechanism of GRU below, and the difference between the GRU and LSTM is also shown in Fig. 5.

The whole mechanism of GRU is summarized in four main parts, shown in Fig. 6. First is the Update gate, it starts with calculating the update gate $z_t$ for time step $t$ by means of equation (2).

$$z_t = \sigma(W^{(z)}x_t + U^{(z)}h_{t-1}) \tag{2}$$

When $x_t$ is applied into the network unit, it is multiplied by its own weight $W^{(z)}$. The same will happen for $h_{t-1}$ which keeps the information for the previous $t-1$ units and multiplied by its own weight $U^{(z)}$. This section adds both results to each other and applies the linear activation function to squash the results. The second main part is the reset gate, and this part is used for the model to decide how much of the past information is forgotten and how much of it is kept by equation (3).

$$r_t = \sigma(W^{(r)}x_t + U^{(r)}h_{t-1}) \tag{3}$$

The difference between parts one and two is weight and gate usage. Now we add the results with their corresponding weights and apply the linear activation function.

The third part concerns the current memory content, and we see the effect of gates on the final output. We start with the use of the reset gate to store the relevant information from the past. It is achieved using equation (4).

$$h'_t = \tanh(Wx_t + r_t \odot Uh_{t-1}) \tag{4}$$

First, we multiply the input $x_t$ with a weight $W$ and $h_{t-1}$ with a weight $U$. In the next step, calculate the Hadamard (element-wise) product between the reset gate $Uh_{t-1}$. That will determine what to eliminate from the previous time steps.

Third, we sum up the results of steps one and two. In the final subset of this main part, we apply the nonlinear activation function (i.e., $tanh$).

The fourth part is the final memory step in which the network needs to calculate the $h'_t$ vector, which keeps information for the current unit and passes it down to the network. For this purpose, the model needs an update gate. It determines what to collect from current memory content $h'_t$ and what from the previous step $h_{t-1}$ the process is:

$$h_t = z_t \odot h_{t-1} + (1 - z_t) \odot h'_t \tag{5}$$

First, apply element-wise multiplication to the update gate $z_t$ and $h_{t-1}$. next is, apply element-wise multiplication to $1 - z_t$ and $h'_t$. Then add the results from the previous two steps.

They are able to store and filter the information using their update and reset gates. So, this model can vanish the gradient problems and perform well. GRUs have fewer tensor operations, and this advantage makes GRUs much faster than LSTM, training-wise.

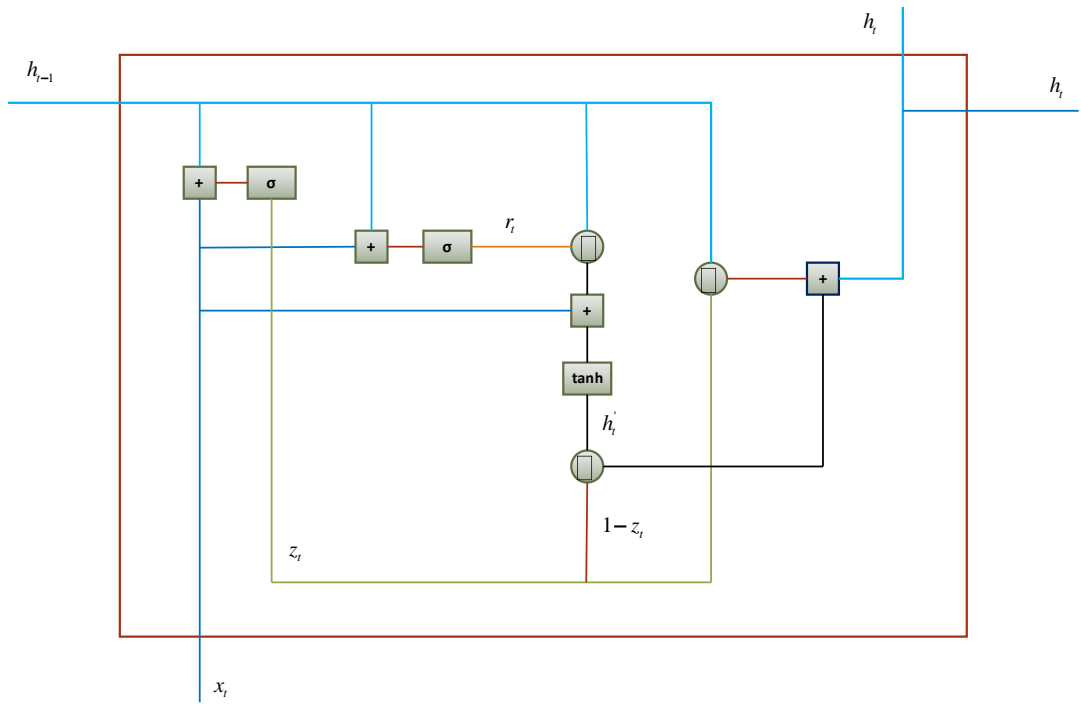

**Figure 6.** GRU mechanism

### 4.3.3 3D CNN

The second branch of our hybrid model and approach is a 3-Dimensional convolutional neural network or C3D. C3D is a deep 3-Dimensional convolution neural network with a homogeneous architecture of $3*3*3$ convolution kernels and $2*2*2$ pooling layers. The advantage of the C3D is that it extracts both spatial and temporal features. Fig.3 shows the schema of a 3-Dimensional convolution model.

The 3D convolution comes from convolving a 3D kernel to the cube formed by stacking multiple contiguous frames together. According to this, the feature maps in the convolution layer are connected to the multiple contiguous frames from the previous layer to get motion information. The below equation shows the value at three $x, y, z$ dimensions, and $jth$ feature maps in the $ith$ layer.

$$v_{ij}^{xyz} = \tanh(b_{ij} + \sum_m \sum_{p=0}^{P_i-1} \sum_{q=0}^{Q_i-1} \sum_{q=0}^{R_i-1} \omega_{ijm}^{pqr} v_{(i-1)m}^{(x+p)(y+q)(z+r)}) \tag{6}$$

Where $R_i$ is the size of the 3D kernel along the temporal connection $ijmpqr$ is the $(p, q, r)$ $th$ value of the kernel connected to the feature map in the previous layer. This method is able to extract only one type of feature from the frame cube since the kernel weights are replicated across the entire cube. To solve this, we increase the number of feature maps in the later layers by implementing the multi-type features from the set of lower-level feature maps [23,24].

Using this method, we extract the spatial and temporal features of the trajectories. The difference between implementing 2D convolution and 3D convolution here is that in 2D convolution, we can only extract spatial features, but this 3D model gives us the advantage of extracting the spatial and temporal features at the same time with better accuracy. In this paper, this method is parallel to the CNN-GRU model, as shown in figure 7.

### 4.3.4 Hybrid CG3D

This section is about concatenating the results of two parallel bodies of our proposed approach and completing the whole model**.** In this section, we get the results from two models of prediction, CNN-GRU, C3D, and concatenate them together, then implement a reshape to show the final result. The final model is a hybrid CG3D model, which has the advantage of high accuracy compared to other trajectory prediction models. The model architecture which demonstrates the whole approach, is shown in Fig 7.

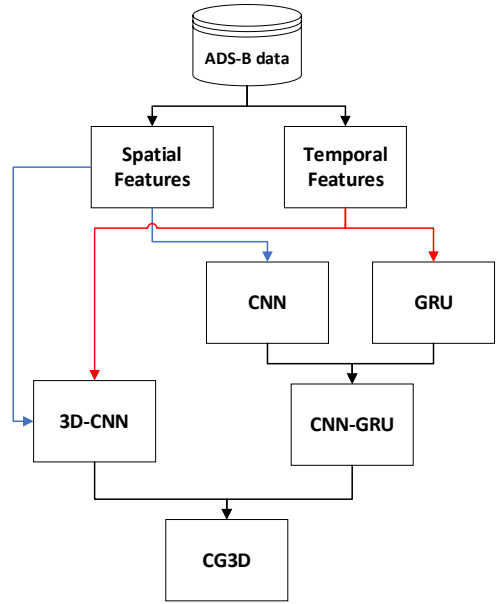

**Figure 7.** CG3D schematic

### 4.4 Network Uncertainty

Air traffic is a complex system with input uncertainties: weather condition changes, inconstant airspeeds, actual take-off mass, and model uncertainties [16]. Uncertainty consideration represents the best model prediction performance. These uncertainties can impact efficiency in operations and, in particular, trajectory prediction. Suppose one of these uncertainties is not taken into account. In that case, it could cause a problematic situation with severe consequences on air traffic [25,26].
There are two types of uncertainty for the modeling process.

1. Aleatoric Uncertainty: In this type, uncertainty is representative of unknowns that differ across natural experiments, this is irreducible in practice.

2. Epistemic Uncertainty: This uncertainty is reducible by adding more knowledge and parameters about the model, gathering more data, etc.

We should focus on the Epistemic type of uncertainty because of its capability to reduce uncertainty. This paper focused on model uncertainty. Implementing the uncertainty in our model makes it more accurate and trustworthy. We consider the Bayesian perspective for our paper uncertainties, the Bayesian perspective of working on uncertainties has currently got the researchers' attention [27,28].

In this paper, we apply a significant approach that can provide uncertainty estimation at prediction time for every neural network, particularly our hybrid model, called Dropout, which is based on the Bayesian algorithm. This method has advantages like preventing overfitting [29,30]. Monte Carlo Dropout (MC Dropout) has a procedure to ignore a random subset of neurons in a network layer at every batch evaluation and running part of the model. In terms of comparison, this method enhances the prediction performance of the proposed model.

We should designate an appropriate level and the number of dropout rates to get the most optimized results for our prediction, which has been affected by uncertainties. MC Dropout has a number of bold advantages, this technique can boost the performance of any trained model, but there is no need to train it again or modify it. MC Dropout is much better in measuring model uncertainty, and it has the important attribute of being simple to implement compared to other models and techniques [31]. We decided to add MC Dropout to our proposed model and compare the training and prediction results in section 5, experiments, table 3.

## 5. Experimental Results

In this section, we demonstrate the computational results and their attachments. Here, we use OpenSky ADS-B data, implemented by means of the Keras TensorFlow backend. Then, we compare the results of the proposed model to the other models using different metrics and plots shown in the following sections. The system we used for training the dataset is a notebook with a Core i7 10750H CPU, RTX2060 GPU, and 16 GB of RAM. For the number of epochs, we choose 500, and the batch size is 512 for all the models.

### 5.1 Performance measurement

Mean Absolute Error (MAE) and Root Mean Squared Error (RMSE) indicators are used for the performance measurement. Each of these metrics is computed with the below equations:

$$MAE = \frac{1}{n}\sum_{t=1}^{n}|F_p - F_t| \qquad (7)$$

$$RMSE = \sqrt{\frac{1}{n}\sum_{t=1}^{n}(F_p - F_t)^2} \qquad (8)$$

Where $F_p$ represents the prediction of our proposed model and each of the comparison models and $F_t$ represents the actual value of the data.

The results are shown in Tables 2 and 3. The proposed CG3D model is more accurate than the other model, and the proposed model with the MC Dropout is more accurate than the one without MC Dropout.

**Table 2.** 4D flight trajectory prediction performance comparison with other models

| Model | MAE | RMSE |
|---|---|---|
| **CG3D** | 0.1776 | 0.2626 |
| **3D CNN** | 0.1785 | 0.2646 |
| **CNN-GRU** | 0.2164 | 0.3728 |

**Table 3.** MCDropout implementation effect on proposed CG3D model

| Model | MAE | RMSE |
|---|---|---|
| **CG3D CNN with MC Dropout** | 0.1406 | 0.2231 |
| **CG3D without MC Dropout** | 0.1776 | 0.2626 |

As shown in Figures 8, a, and b, the error measurement results indicate the higher accuracy and better performance of the CG3D model than other models. The error measurements RMSE and MAE for the proposed model vary within a larger range than other models, but the model has a lower average error value. Figures 8, c, and d represent the MSE loss and MSE validation loss for each model. The values in d are lower than those in c for all models, which suggests that overfitting is not a problem. Again the CG3D CNN with MC Dropout has the best performance.

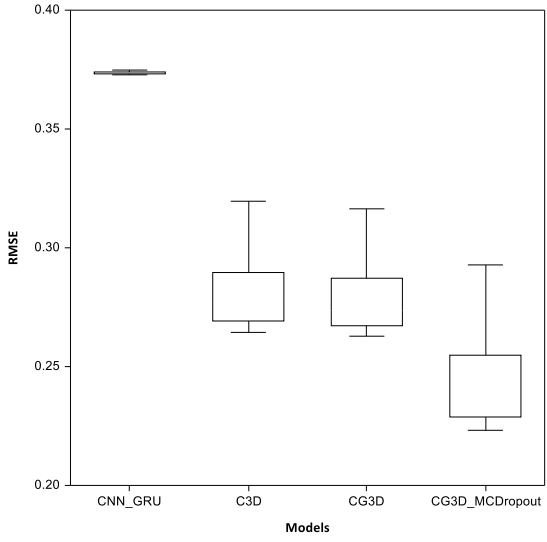 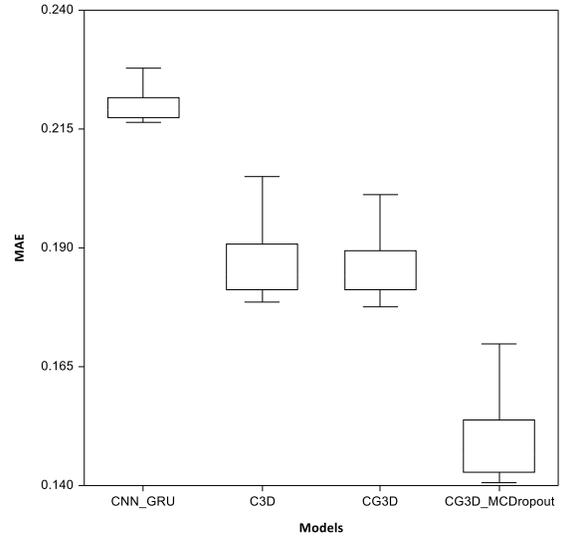

(a) (b)

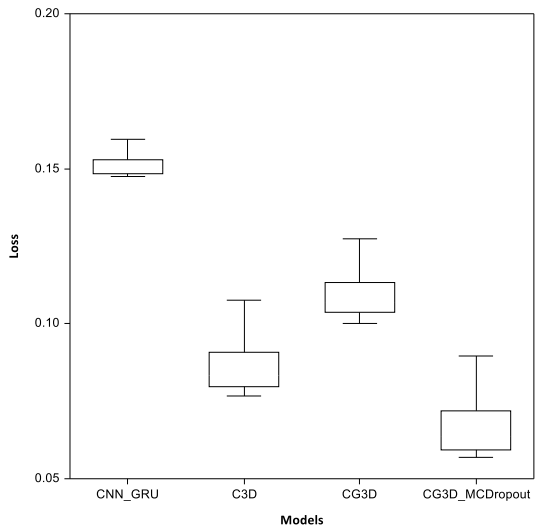 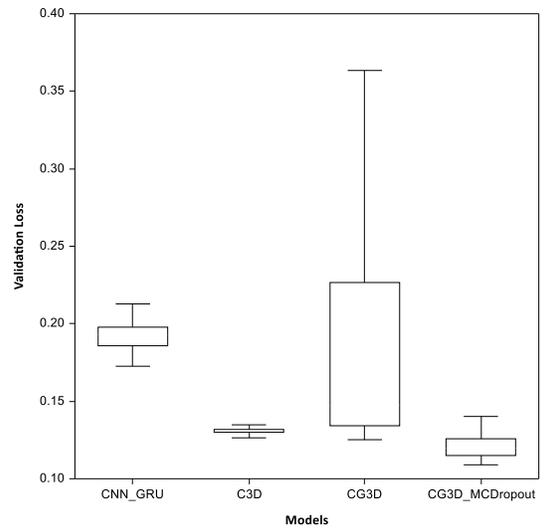

(c) (d)

**Figure 8.** Error measurement results; (a) RMSE for models; (b) MAE for models; (c) MSE loss; (d) MSE validation loss

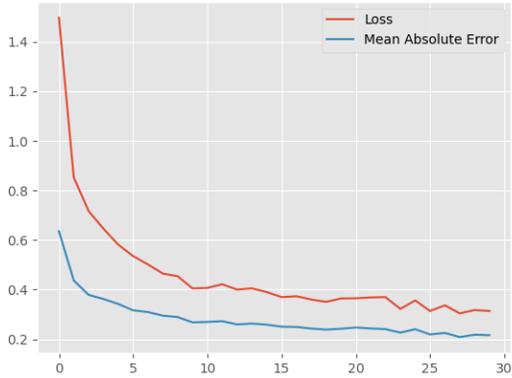 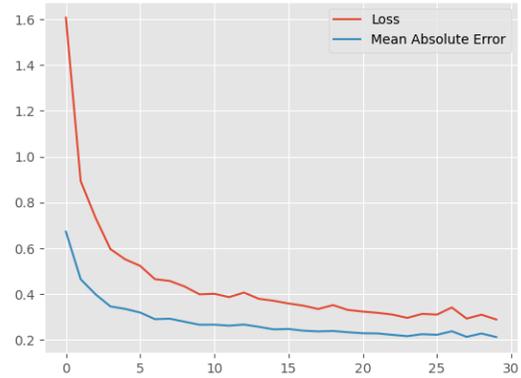

**Figure 9.** Loss and Mean Absolute Error for CG3D without MC Dropout

**Figure 10.** Loss and Mean Absolute Error for CG3D with MC Dropout

Figures 9 and 10 show decreasing loss and validation loss for the proposed model with MC dropouts without MC dropouts consecutively. This plot either demonstrates that the proposed model with the MC Dropout is reached an average 21 % lower error value for the 4D flight trajectory prediction.

## 6. Conclusion and Future Research

This paper proposes a hybrid deep learning model for 4D flight trajectory prediction. The model combines a CNN-GRU and a 3D-CNN model for spatial-temporal feature extraction and prediction. The proposed model is implemented for the Hartsfield–Jackson Atlanta International Airport (ATL) based on the historical Automatic Dependent Surveillance-Broadcast (ADS–B) dataset. This model has the advantages of higher prediction accuracy, long-term prediction, and sufficient spatial-temporal feature extraction compared to the other prediction models (i.e., CNN-GRU, 3D CNN) used in this research. MAE and RMSE indicators are used as objective error functions to evaluate the performance of the 4D trajectory prediction models. The prediction performance of the proposed model is compared with CNN-GRU, 3DCNN, and the proposed model without the MC Dropout using these error measurements. The experimental results show the proposed model performs well with the MC dropout compared to other deep models.

Efficient and robust strategical flight schedules wil help airports, airlines, and also passengers. The proposed model can precisely predict the 4D flight trajectory. The error values decreased by an average of 20-35% compared to the other models, demonstrating the improvement of the proposed model.

The following steps need to be further investigated in future studies. Preprocessing steps can be taken into account for eliminating data anomalies such as missing and duplicated data. The performance of the deep predictor model can be compared based on different input data determination strategies. The input to the model can either be a unified dataset, and a separate dataset consists of Spatio-temporal features. This paper uses the novel separated Spatio-temporal input approach. We compare the prediction with our novel separated Spatio-temporal approach to the those with unified input. Finally, consideration of different sources of uncertainty in the input data can be evaluated in the future.

## 7. Acknowledgment

The authors would like to thank the OpenSky-network website for providing open access to their vast and complete ADS-B dataset. The data is available at https://opensky-network.org/.## 8. References

[1] Statistica. (2021, April 22). Passenger traffic at Atlanta Airport 2000-2020. https://www.statistica.com/statistics/226464/passenger-traffic-at-atlanta-airport/

[2] G. Luckenbaugh, S. Landriau, J. Dehn, and S. Rudolph, "Service oriented architecture for the next generation air transportation system," in Proc. Integr. Commun., Navigat. Surveill. Conf., Herndon, VA, USA, 2007, pp. 1–9.

[3] Harrison, Michael J. "ADS-X the next gen approach for the next generation air transportation system." *2006 ieee/aiaa 25TH Digital Avionics Systems Conference*. IEEE, 2006.

[4] Sipe, Alvin, and John Moore. "Air traffic functions in the NextGen and SESAR airspace." *2009 IEEE/AIAA 28th Digital Avionics Systems Conference*. IEEE, 2009.

[5] Guan, Xiangmin, et al. "A strategic flight conflict avoidance approach based on a memetic algorithm." *Chinese Journal of Aeronautics* 27.1 (2014): 93-101.

[6] J. Li, Q. Li, N. Chen, and Y. Wang, "Indoor pedestrian trajectory detection with LSTM network," in Proc. 7 IEEE Int. Conf. Comput. Sci. Eng. (CSE), Guangzhou, China, Jul. 2017, pp. 651–654.

[7] Y. Liu, H. Zheng, X. Feng, and Z. Chen, "Short-term traffic flow prediction with conv-LSTM," in Proc. 9th Int. Conf. Wireless Commun. Signal Process. (WCSP), Nanjing, China, Oct. 2017, pp. 1–6.

[8] Zhang, Junfeng, JieLiu, RongHU, Haibo Zhu "Online four dimensional trajectory prediction method based on aircraft intent updating." *Aerospace Science and Technology* 77 (2018): 774-787.

[9] Zhang, Xiaoge, and Sankaran Mahadevan. "Bayesian neural networks for flight trajectory prediction and safety assessment." *Decision Support Systems* 131 (2020): 113246.

[10] Wang, Zhengyi, Man Liang, and Daniel Delahaye. "A hybrid machine learning model for short-term estimated time of arrival prediction in terminal maneuvering area." *Transportation Research Part C: Emerging Technologies* 95 (2018): 280-294.

[11] Ma, Lan, and Shan Tian. "A Hybrid CNN-LSTM Model for Aircraft 4D Trajectory Prediction." *IEEE Access* 8 (2020): 134668-134680.

[12] Guo, Jingqiu, et al. "GPS-based citywide traffic congestion forecasting using CNN-RNN and C3D hybrid model." *Transportmetrica A: Transport Science* (2020): 1-22.

[13] Lambelho, Miguel, et al. "Assessing strategic flight schedules at an airport using machine learning-based flight delay and cancellation predictions." *Journal of Air Transport Management* 82 (2020): 101737.15

[14] Shi, Zhiyuan, Min Xu, and Quan Pan. "4-D Flight Trajectory Prediction with Constrained LSTM Network." *IEEE Transactions on Intelligent Transportation Systems* (2020).

[15] Wu, Zhi-Jun, Shan Tian, and Lan Ma. "A 4D Trajectory Prediction Model Based on the BP Neural Network." *Journal of Intelligent Systems* 29.1 (2019): 1545-1557.

[16] Zeh, Thomas, Judith Rosenow, and Hartmut Fricke. "Interdependent Uncertainty Handling in Trajectory Prediction." *Aerospace* 6.2 (2019): 15.

[17] Wang, Zhengyi, Man Liang, and Daniel Delahaye. "Short-term 4d trajectory prediction using machine learning methods." *Proc. SID*. 2017.

[18] Wang, Jing, Yunkai Zou, and Jianli Ding. "ADS-B spoofing attack detection method based on LSTM." EURASIP Journal on Wireless Communications and Networking 2020.1 (2020): 1-12.

[19] T. Alipourfard, H. Arefi, and S. Mahmoudi, "A novel deep learning framework by combination of subspace-based feature extraction and convolutional neural networks for hyperspectral images classification," in Proc. IEEE Int. Geosci. Remote Sens. Symp., Valencia, Spain, Jul. 2018, pp. 4780–4783.

[20] A. T. Vo, H. S. Tran, and T. H. Le, "Advertisement image classification using convolutional neural network," in Proc. 9th Int. Conf. Knowl. Syst. Eng. (KSE), Oct. 2017, pp. 197–202.

[21] Z. Peng, Y. Tao, L. Yanan, F. Zhiyong, and D. Zhaobin, "Feature extraction and prediction of QAR data based on CNN-LSTM," Appl. Res. Comput., vol. 36, no. 10, pp. 2958–2961, 2019.

[22] Michele Phi 2018, towards data science, accessed January 23 2021, <https://towardsdatascience.com/illustrated-guide-to-lstms-and-gru-s-a-step-by-step-explanation-44e9eb85bf21>

[23] Ji, Shuiwang, et al. "3D convolutional neural networks for human action recognition." IEEE transactions on pattern analysis and machine intelligence 35.1 (2012): 221-231.

[24] Guo, Jingqiu, et al. "Deep learning based congestion prediction using PROBE trajectory data." *CICTP 2019*. 2019. 3136-3147.

[25] Rivas, Damián, Antonio Franco, and Alfonso Valenzuela. "Analysis of aircraft trajectory uncertainty using Ensemble Weather Forecasts." *Proc. 7th European Conference for Aeronautics and Space Sciences (EUCASS)*. 2017.

[26] Srivastava, Nitish, et al. "Dropout: a simple way to prevent neural networks from overfitting." *The journal of machine learning research* 15.1 (2014): 1929-1958.

[27] Wang, Sida, and Christopher Manning. "Fast dropout training." *international conference on machine learning*. PMLR, 2013.

[28] Maeda, Shin-ichi. "A Bayesian encourages dropout." *arXiv preprint arXiv:1412.7003* (2014).16